\documentclass[numbers]{article}

\usepackage[preprint]{neurips_2022}

\usepackage[utf8]{inputenc} 
\usepackage[T1]{fontenc}    
\usepackage{hyperref}       
\usepackage{url}            
\usepackage{booktabs}       
\usepackage{amsfonts}       
\usepackage{nicefrac}       
\usepackage{microtype}      
\usepackage{xcolor}         
\usepackage{graphicx}
\usepackage{amsmath}
\usepackage{stmaryrd}

\title{A Survey on Offline Model-Based Reinforcement Learning}

\author{%
  Haoyang He \\
  Electrical and Computer Engineering \\
  Carnegie Mellon University\\
  Pittsburgh, PA 15213 \\
  \texttt{hhe2@andrew.cmu.edu} \\
}

\begin{document}

\maketitle

\begin{abstract}
    Model-based approaches are becoming increasingly popular in the field of offline reinforcement learning, with high potential in real-world applications due to the model's capability of thoroughly utilizing the large historical datasets available with supervised learning techniques. This paper presents a literature review of recent work in offline model-based reinforcement learning, a field that utilizes model-based approaches in offline reinforcement learning. The survey provides a brief overview of the concepts and recent developments in both offline reinforcement learning and model-based reinforcement learning, and discuss the intersection of the two fields. We then presents key relevant papers in the field of offline model-based reinforcement learning and discuss their methods, particularly their approaches in solving the issue of distributional shift, the main problem faced by all current offline model-based reinforcement learning methods. We further discuss key challenges faced by the field, and suggest possible directions for future work.
\end{abstract}

\section{Introduction}
Reinforcement learning (RL) constitutes a subfield of machine learning, concentrating on the development of agents capable of making optimal decisions within uncertain environments. In an RL setting, agents engage with their surroundings by performing actions and receiving feedback in the form of rewards or penalties. The primary objective is for the agent to ascertain a policy, a mapping from states to actions, which maximizes the accumulated reward over a specified time period \cite{sutton2018reinforcement}.

\subsection{Offline Reinforcement Learning}
Offline reinforcement learning, also referred to as batch reinforcement learning, entails a form of reinforcement learning wherein the policy is exclusively learned based on pre-existing historical data. In practical applications, the available historical dataset is typically extensive, and offline reinforcement learning seeks to maximize the utilization of this data, learning the policy as a decision-making engine, contrasting with online reinforcement learning's emphasis on exploration \cite{10078377}. In practice, the preceding data collection policy, responsible for generating the historical data, is customarily the current method or model in deployment. Recent research endeavors often concentrate on optimizing the algorithm's outcome in relation to the data collection policy, with the aim of devising new algorithms possessing practical deployment value \cite{bhardwaj2023adversarial, bremen}. The algorithm grapples with the challenge of distributional shift, signifying that the deployment scenario exhibits a distinct distribution in comparison to the historical dataset, with respect to state and action space \cite{levine2020offline}. The presence of distributional shift constitutes a key obstacle in offline reinforcement learning, warranting further in-depth examination.

\subsection{Model-Based Reinforcement Learning}
Model-based reinforcement learning encompasses a category of reinforcement learning methodologies that focus on constructing a model to comprehend the environment by approximating state transition dynamics and reward functions. By leveraging the estimated model, these algorithms typically generate simulated data to facilitate the learning of an optimal policy, in conjunction with the actual data, also known as experienced data. Enhancing the performance of the learned policy by capitalizing on the model necessitates acquiring a model that more precisely characterizes the actual environment \cite{luo2022survey}. Numerous techniques for augmenting model learning have been investigated in the literature over time, including the use of approximation functions trained with prediction loss, which is susceptible to horizontal compounding error during updates. In order to mitigate this error, research has proposed implementing constraints on the model, such as the Lipschitz Continuity Constraint \cite{asadi2018lipschitz}. An alternative approach to diminishing this error involves distribution matching via adversarial learning against a discriminator \cite{xu2021error}.

\section{Offline Model-Based Reinforcement Learning}
The intersection of the two previously discussed concepts defines offline model-based reinforcement learning. Analogous to online model-based RL methods, offline model-based RL approaches initially estimate state transition dynamics and reward functions, subsequently employing them for environment simulation and planning based on both simulated data and experienced data. Owing to the characteristics of model-based learning, offline model-based learning may particularly benefit from the robust model established by supervised learning, which takes full advantage of the extensive historical dataset when compared to offline model-free methods \cite{10078377}. Nevertheless, due to the intrinsic properties of offline learning, offline model-based RL also grapples with distributional shifts in both state and action distributions. Unlike online methods, no exploration is available for the model to rectify itself. This challenge is exacerbated by model exploitation since the model is derived from data collected by another policy with a distributional shift relative to the deployment environment. Consequently, simple policy optimization over the model may result in an optimized outcome in the historical distribution, which is out of distribution for the deployment environment, potentially leading to inferior results. Recent advancements in offline model-based reinforcement learning have concentrated on addressing the issue of distributional shift, typically by imposing certain constraints on the model through modifications in state transition dynamics, reward functions, or value functions \cite{li2022settling, morel, mopo, bremen, combo, rigter2022ramborl, bhardwaj2023adversarial}. 

\subsection{MOReL}
Kidambi et al. proposed MOReL \cite{morel}, a framework for model-based offline reinforcement learning that penalizes the reward for uncertainty, allowing the use of classical planning methods with altered reward function and state transition dynamics to address the issue of uncertainty and distributional shift \cite{levine2020offline}. They constructed an Unknown State Action Detector based on a threshold over the total variational divergence between the learned model and the behavior model, which is then used to determine if a certain policy goes into the absorbing state HALT to act as a control for the policy to enter the above-threshold uncertain state, whose reward is set to a certain negative value, addressing the model exploitation problem. In their practical implementation, they constructs an ensemble of state transition dynamics models for uncertainty estimation, using a measurement of disagreement between the models given by eq.\ref{morel dist},
\begin{equation}
    dist_{i, j}(s_t, a_t) = \mathbb{E}_{s_{t+1}^i \sim T_{\psi_i}(\cdot | s_t, a_t), s_{t+1}^i \sim T_{\psi_i}(\cdot | s_t, a_t)}[||s^i_{t+1} - s^j_{t+1}||].
    \label{morel dist}
\end{equation}
in order to define the uncertainty of the state action pair as eq.\ref{morel u}.
\begin{equation}
    U_r(s, a) = \left\{
    \begin{aligned}
        r_{max}, \max_{i, j}dist_{i, j}(s, a) > threshold, \\
        0, \max_{i, j}dist_{i, j}(s, a) \leq threshold.
    \end{aligned}
    \right.
    \label{morel u}
\end{equation}

\subsection{MOPO}
Yu et al. proposed MOPO \cite{mopo}, model-based offline policy optimization, a similar attempt to address the problem of distributional shift by penalizing the reward. In the practical implementation of MOPO, the ensemble of models are estimated using the multivariate Gaussian distributions modeling the $i$th transition dynamics of the model ensembles, denotes in eq.\ref{mopo t},
\begin{equation}
    T_{\psi_i}(s'|s, a) = \mathcal{N}(\mu_i(s, a), \Sigma_i(s, a)).
    \label{mopo t}
\end{equation}
while the uncertainty is estimated by the Frobenius norm of the maximum distributional variance in among the ensemble of models, as shown in eq.\ref{mopo u}.
\begin{equation}
    U_r(s, a) = \max_i||\Sigma_i(s, a)||_F.
    \label{mopo u}
\end{equation}
Both methods are designed to build upon previously existing online model-based RL methods, providing a penalized reward function, which can in turn be directly used instead of the original reward function in classical planning methods \cite{10078377} like linear-quadratic regulator \cite{tassa2012synthesis} or Monte Carlo tree search \cite{browne2012survey}. 

\subsection{BREMEN}
An alternative approach to the distributional shift problem proposed by Matsushima et al. is BREMEN \cite{bremen}, behavior-regularized model ensemble. Instead of constructing a pessimistic or reward-penalized MDP model, BREMEN learns an ensemble of dynamic models in conjunction with a policy using imaginary rollouts from the ensemble and behavior regularization via conservative trust-region updates. Each model of the ensemble is trained to optimize the objective of eq.\ref{bremen_1}, the mean squared error between the prediction of the next state and the true next state, and imaginary model rollouts are generated sequentially.
\begin{equation}
    \min_{\phi_i} \frac{1}{|\mathcal{D}|} \sum_{(s_t, a_t, s_{t+1}) \in \mathcal{D}}\frac{1}{2} ||s_{t+1} - \hat{f}_{\phi_i}(s_t, a_t)||^2_2
    \label{bremen_1}
\end{equation}
To address the distributional shift problem, BREMEN uses an iterative policy update via a trust-region constraint, re-initialized with the behavior cloned policy after every deployment, with the behavior cloning objective function of eq.\ref{bremen_2}.
\begin{equation}
    \min_{\beta} \frac{1}{|\mathcal{D}|} \sum_{(s_t, a_t) \in \mathcal{D}}\frac{1}{2} ||a_t - \hat{\pi}_{\beta}(s_t)||^2_2
    \label{bremen_2}
\end{equation}
After obtaining the estimated behavior policy, the target policy $\pi_\theta$ is initialized as a Gaussian policy with mean from the behavior policy and standard deviation of $1$. This behavior cloning initialization may be seen as implicitly biasing the optimized target policy to be close to the data-collection policy, and thus reduces the distributional shift. To further bias the learned policy to be close to the data-collection policy, KL-divergence trust region under a certain threshold is added as an additional constraint, yielding the optimization problem of eq.\ref{bremen_3}.
\begin{equation}
    \begin{aligned}
        & \theta_{k+1}=\underset{\theta}{\arg \max } \underset{s, a \sim \pi_{\theta_k}, \hat{f}_{\phi_i}}{\mathbb{E}}\left[\frac{\pi_\theta(a \mid s)}{\pi_{\theta_k}(a \mid s)} A^{\pi_{\theta_k}}(s, a)\right], \\
        & \text { s.t. } \quad \underset{s, a \sim \pi_{\theta_k}, \hat{f}_{\phi_i}}{\mathbb{E}} \left[D_{\mathrm{KL}}\left(\pi_\theta(\cdot \mid s) \| \pi_{\theta_k}(\cdot \mid s)\right)\right] \leq \delta, \quad \pi_{\theta_0}=\operatorname{Normal}\left(\hat{\pi}_\beta, 1\right). \\
    \end{aligned}
    \label{bremen_3}
\end{equation}
The authors claim an empirical improvement over direct KL-divergence penalties such as previous methods, and showed that KL-divergence is controlled implicitly in the optimization. The author also highlighted that BREMEN may outperform similar methods in offline settings with limited data, attributing to its sequential imaginary behavioral model rollouts, hence having higher sample efficiency. 

\subsection{VI-LCB}
VI-LCB \cite{vilcb}, offline value iteration with lower confidence bound, is an alternative approach to addressing the uncertainty created by distributional shift proposed by Rashidinejad et al. With a similar idea of penalizing state-action pairs under poor or partial coverage, VI-LCB builds upon the classic value iteration algorithm, and instead of iterating upon the original values given by the learned model, VI-LCB employs the concept of pessimism \cite{vilcb} and subtracts a penalty function to the value function. The value function for value iteration in VI-LCB is defined by eq.\ref{VI-LCB q}, \ref{VI-LCB v}.
\begin{equation}
    Q(s, a) \mapsfrom r(s, a) - b(s, a) + \gamma \sum_{s'} P(s'|s, a)V(s'), \forall (s, a),
    \label{VI-LCB q}
\end{equation}
\begin{equation}
    V(s) \mapsfrom \max_a Q(s, a), \forall s.
    \label{VI-LCB v}
\end{equation}
VI-LCB is developed shortly after PEVI \cite{pmlr-v139-jin21e}, pessimistic value iteration  proposed by Jin et al. It is a simple framework to utilize pessimism in offline reinforcement learning, which simply changes the addition to subtraction for the term encouraging exploration in online RL algorithms to use the term to penalize less-explored state-action pairs.

Li et al. proposed a version of VI-LCB with Bernstein-style penalties, denoted by eq.\ref{bernstein}. 
\begin{equation}
    b(s, a; V) := \min \{ \max \{ \sqrt{\frac{c_b \log \frac{N}{(1-\gamma)\delta}}{N(s, a)} Var_{\hat{P}_{s, a}} (V)}, \frac{2c_b \log \frac{N}{(1-\gamma)\delta}}{(1-\gamma)N(s, a)} \}, \frac{1}{1-\gamma}\} + \frac{5}{N}.
    \label{bernstein}
\end{equation}
Based on this version of VI-LCB, Li et al. further proved that model-based approaches in offline reinforcement learning achieve minimax-optimal sample complexity without burn-in cost for tabular Markov Decision Process (MDP) \cite{li2022settling}. Specifically, given a single-policy clipped concentrability coefficient $C^*_{clipped}$, model-based offline RL yields $\epsilon$-accuracy with sample complexity given by
\begin{equation}
    \left\{
    \begin{aligned}
        \frac{SC^*_{clipped}}{(1-\gamma)^3 \epsilon^2}, \text{infinite-horizon MDPs}, \\
        \frac{H^4SC^*_{clipped}}{\epsilon^2}, \text{finite-horizon MDPs}.
    \end{aligned}
    \right.
\end{equation}
for $\gamma$-discounted infinite-horizon MDPs with effective horizon $\frac{1}{1-\gamma}$ or finite-horizon MDPs with horizon $H$.

\subsection{COMBO}
COMBO \cite{combo}, conservative offline model-based policy optimization, is an approach proposed by Yu et al. to prevent distributional shift by directly optimizing the model learning without explicitly providing an uncertainty estimation in the planning process. In the construction of the model rollouts, each policy is repeatedly evaluated on the conservative policy estimation function \ref{combo_2},
\begin{equation}
    \hat{Q}^{k+1} \leftarrow \arg \min _Q \beta\left(\mathbb{E}_{\mathbf{s}, \mathbf{a} \sim \rho(\mathbf{s}, \mathbf{a})}[Q(\mathbf{s}, \mathbf{a})]-\mathbb{E}_{\mathbf{s}, \mathbf{a} \sim \mathcal{D}}[Q(\mathbf{s}, \mathbf{a})]\right)+\frac{1}{2} \mathbb{E}_{\mathbf{s}, \mathbf{a}, \mathbf{s}^{\prime} \sim d_f}\left[\left(Q(\mathbf{s}, \mathbf{a})-\widehat{\mathcal{B}}^\pi \hat{Q}^k(\mathbf{s}, \mathbf{a})\right)^2\right].
    \label{combo_2}
\end{equation}
which penalizes rewards that are more likely to be out of distribution, based on the sampling distribution, a hyperparameter of choice to determine the desired conservativeness. The policy is then improved using the conservative critic Q obtained from previous conservative estimation \ref{combo_3}. 
\begin{equation}
    \hat{\pi} \leftarrow \arg \max_\pi \mathbb{E}_{s\sim \rho, a \sim \pi(\cdot | s)}\left[ \hat{Q}^\pi(\mathbf{s}, \mathbf{a})\right].
    \label{combo_3}
\end{equation}
The authors also showed that the obtained policy has a safe improvement guarantee over the behavior policy. A key improvement of COMBO over similar previous methods like MOReL and MOPO is that it no longer requires an uncertainty estimation in order to perform optimization, which can in itself be inaccurate and hard to obtain \cite{10078377}, but instead utilizes an adversarial training process to optimize itself. 

\subsection{RAMBO}
Rigter et al. proposed RAMBO-RL \cite{rigter2022ramborl}, Robust Adversarial Model-Based Offline Reinforcement Learning, a self-adversarial model. Instead of previous adversarial RL models that train an adversary, RAMBO directly trains the model adversarially, while enforcing the conservatism in the training of the adversarial dynamics model. The algorithm is basically an iterative update of the optimization of the adversarial model updating the state transition dynamics, where the optimization objective of the adversarial model is based on a formulation of optimization objective to reduce distributional shift of a previous work Uehara et al. \cite{uehara2021pessimistic}, modified to eq.\ref{rambo_6},
\begin{equation}
    \min _{\widehat{T}_\phi} V_\phi^\pi, \quad \text { s.t. } \mathbb{E}_{\mathcal{D}}\left[\mathrm{TV}\left(\widehat{T}_{\mathrm{MLE}}(\cdot \mid s, a), \widehat{T}_\phi(\cdot \mid s, a)\right)^2\right] \leq \xi,
    \label{rambo_6}
\end{equation}
which can be rewritten with Lagrangian relaxation to eq.\ref{rambo_7}.
\begin{equation}
    \max _{\lambda \geq 0} \min _{\widehat{T}_\phi}\left(L(\hat{T}, \lambda):=V_\phi^\pi+\lambda\left(\mathbb{E}_{\mathcal{D}}\left[\mathrm{TV}\left(\widehat{T}_{\mathrm{MLE}}(\cdot \mid s, a), \widehat{T}_\phi(\cdot \mid s, a)\right)^2\right]-\xi\right)\right).
    \label{rambo_7}
\end{equation}
Rather than optimizing the Lagrangian multiplier, the authors claimed it was easier to fix the multiplier as a hyperparameter for tuning, reducing it as a scaling factor, which can then be reduced to eq.\ref{rambo_8}.
\begin{equation}
    \min _{\widehat{T}_\phi}\left(\lambda V_\phi^\pi+\mathbb{E}_{\mathcal{D}}\left[\operatorname{TV}\left(\widehat{T}_{\mathrm{MLE}}(\cdot \mid s, a), \widehat{T}_\phi(\cdot \mid s, a)\right)^2\right]\right).
    \label{rambo_8}
\end{equation}
To make the optimization more simple, the authors changed from optimizing the total variational divergence to simple MLE loss, which leads to the final optimization objective eq.\ref{rambo_9}.
\begin{equation}
    \mathcal{L}_\phi=\lambda V_\phi^\pi-\mathbb{E}_{\left(s, a, r, s^{\prime}\right) \sim \mathcal{D}}\left[\log \widehat{T}_\phi\left(s^{\prime}, r \mid s, a\right)\right].
    \label{rambo_9}
\end{equation}
The authors explicitly compared RAMBO to COMBO, and claimed that the pessimistic value function updates used by COMBO create local maxima in the Q-function which are present throughout training, making COMBO likely to stuck in a local maxima, whereas RAMBO’s adversarial modifications resulted in a less likelihood. 

\subsection{ARMOR}
Bhardwaj et al. proposed ARMOR \cite{bhardwaj2023adversarial}, adversarial model for offline reinforcement learning, a novel framework capable of improving upon an arbitrary reference policy regardless of data quality.  The two-player game consists of a learner policy and an adversary MDP model, defined as eq.\ref{armor}.
\begin{equation}
    \hat{\pi} = \arg \max_{\pi \in \Pi} \min_{M \in \mathcal{M}_\alpha} J_M(\pi) - J_M(\pi_{ref}).
    \label{armor}
\end{equation}
In the algorithm, the adversary MDP model is updated in the process \ref{armor adv} for $i = 1, 2$, where $\mathcal{D}_M$ is the transition tuples using model predictions based on minibatches of dataset samples from real and model data, $f_1, f_2$ are critic networks used by the double Q residual algorithm loss from \cite{cheng2022adversarially}. 
\begin{equation}
    \begin{aligned}
        l^{\text {adversary }}(f, M) & :=\mathcal{L}_{\mathcal{D}_M}\left(f, \pi, \pi_{\text {ref }}\right)+\beta\left(\mathcal{E}_{\mathcal{D}_M}^w(f, M, \pi)+\lambda \mathcal{E}_{\mathcal{D}_{\text {real }}^{\text {min }}}(M)\right), \\
        M & \leftarrow M-\eta_{\text {fast }}\left(\nabla_M l^{\text {adversary }}\left(f_1, M\right)+\nabla_M l^{\text {adversary }}\left(f_2, M\right)\right), \\
        f_i & \leftarrow \operatorname{Proj}_{\mathcal{F}}\left(f_i-\eta_{\text {fast }} \nabla_{f_i} l^{\text {adversary }}\left(f_i, M\right)\right), \\
        \bar{f}_i & \leftarrow(1-\tau) \bar{f}_i+\tau f_i.
    \end{aligned}
    \label{armor adv}
\end{equation}
Then the actor network is updated with respect to the first critic $f_1$ and the reference policy in the process \ref{armor act}.
\begin{equation}
    \begin{aligned}
        l^{\text {actor }}(\pi) & :=-\mathcal{L}_{\mathcal{D}_M}\left(f_1, \pi, \pi_{\text {ref }}\right), \\
        \pi & \leftarrow \operatorname{Proj}_{\Pi}\left(\pi-\eta_{\text {slow }} \nabla_\pi l^{\text {actor }}(\pi)\right).
    \end{aligned}
    \label{armor act}
\end{equation}
The model state is updated by query of the MDP model after the updates to the two players. Inspired by the idea of relative pessimism proposed by Cheng et al. \cite{cheng2022adversarially}, ARMOR optimizes for the worst-case relative performance over uncertainty, instead of a predetermined penalty term based on the state-action itself. The authors showed that with relative pessimism, ARMOR policy does not degrades the performance of a reference policy when given a fixed set of hyperparameters, and is capable of competing against any policy covered by the data when the right hyperparameter is chosen.

\section{Challenges and Future Work}
Distributional shift remains the most influential problem in offline model-based reinforcement learning, and warrants further research. Even though the discussed relevant work have proposed various methods to mitigate this problem, none of the approaches have succeeded in providing a theoretically provable data non-reliant absolute performance guarantee due to this problem. 

Another promising direction is the absolute relative performance of offline model-based reinforcement learning. As highlighted by \cite{bhardwaj2023adversarial}, many theoretically proven methods are not deployed in real-world scenarios because they failed to achieve absolute more optimal performance over the current policy, making policy replacement infeasible. Future work on the absolute relative performance of learned policies can greatly improve the deployment of novel offline model-based reinforcement learning algorithms.

\bibliographystyle{plain}
\bibliography{references.bib}

\begin{thebibliography}{10}

\bibitem{asadi2018lipschitz}
Kavosh Asadi, Dipendra Misra, and Michael Littman.
\newblock Lipschitz continuity in model-based reinforcement learning.
\newblock In {\em International Conference on Machine Learning}, pages
  264--273. PMLR, 2018.

\bibitem{bhardwaj2023adversarial}
Mohak Bhardwaj, Tengyang Xie, Byron Boots, Nan Jiang, and Ching-An Cheng.
\newblock Adversarial model for offline reinforcement learning, 2023.

\bibitem{browne2012survey}
Cameron~B Browne, Edward Powley, Daniel Whitehouse, Simon~M Lucas, Peter~I
  Cowling, Philipp Rohlfshagen, Stephen Tavener, Diego Perez, Spyridon
  Samothrakis, and Simon Colton.
\newblock A survey of monte carlo tree search methods.
\newblock {\em IEEE Transactions on Computational Intelligence and AI in
  games}, 4(1):1--43, 2012.

\bibitem{cheng2022adversarially}
Ching-An Cheng, Tengyang Xie, Nan Jiang, and Alekh Agarwal.
\newblock Adversarially trained actor critic for offline reinforcement
  learning.
\newblock In {\em International Conference on Machine Learning}, pages
  3852--3878. PMLR, 2022.

\bibitem{pmlr-v139-jin21e}
Ying Jin, Zhuoran Yang, and Zhaoran Wang.
\newblock Is pessimism provably efficient for offline rl?
\newblock In Marina Meila and Tong Zhang, editors, {\em Proceedings of the 38th
  International Conference on Machine Learning}, volume 139 of {\em Proceedings
  of Machine Learning Research}, pages 5084--5096. PMLR, 18--24 Jul 2021.

\bibitem{morel}
Rahul Kidambi, Aravind Rajeswaran, Praneeth Netrapalli, and Thorsten Joachims.
\newblock Morel: Model-based offline reinforcement learning.
\newblock In H.~Larochelle, M.~Ranzato, R.~Hadsell, M.F. Balcan, and H.~Lin,
  editors, {\em Advances in Neural Information Processing Systems}, volume~33,
  pages 21810--21823. Curran Associates, Inc., 2020.

\bibitem{levine2020offline}
Sergey Levine, Aviral Kumar, George Tucker, and Justin Fu.
\newblock Offline reinforcement learning: Tutorial, review, and perspectives on
  open problems, 2020.

\bibitem{li2022settling}
Gen Li, Laixi Shi, Yuxin Chen, Yuejie Chi, and Yuting Wei.
\newblock Settling the sample complexity of model-based offline reinforcement
  learning.
\newblock {\em arXiv preprint arXiv:2204.05275}, 2022.

\bibitem{luo2022survey}
Fan-Ming Luo, Tian Xu, Hang Lai, Xiong-Hui Chen, Weinan Zhang, and Yang Yu.
\newblock A survey on model-based reinforcement learning, 2022.

\bibitem{bremen}
Tatsuya Matsushima, Hiroki Furuta, Yutaka Matsuo, Ofir Nachum, and Shixiang Gu.
\newblock Deployment-efficient reinforcement learning via model-based offline
  optimization, 2020.

\bibitem{10078377}
Rafael~Figueiredo Prudencio, Marcos R. O.~A. Maximo, and Esther~Luna Colombini.
\newblock A survey on offline reinforcement learning: Taxonomy, review, and
  open problems.
\newblock {\em IEEE Transactions on Neural Networks and Learning Systems},
  pages 1--0, 2023.

\bibitem{vilcb}
Paria Rashidinejad, Banghua Zhu, Cong Ma, Jiantao Jiao, and Stuart Russell.
\newblock Bridging offline reinforcement learning and imitation learning: A
  tale of pessimism.
\newblock In M.~Ranzato, A.~Beygelzimer, Y.~Dauphin, P.S. Liang, and J.~Wortman
  Vaughan, editors, {\em Advances in Neural Information Processing Systems},
  volume~34, pages 11702--11716. Curran Associates, Inc., 2021.

\bibitem{rigter2022ramborl}
Marc Rigter, Bruno Lacerda, and Nick Hawes.
\newblock Rambo-rl: Robust adversarial model-based offline reinforcement
  learning, 2022.

\bibitem{sutton2018reinforcement}
Richard~S Sutton and Andrew~G Barto.
\newblock {\em Reinforcement learning: An introduction}.
\newblock MIT press, 2018.

\bibitem{tassa2012synthesis}
Yuval Tassa, Tom Erez, and Emanuel Todorov.
\newblock Synthesis and stabilization of complex behaviors through online
  trajectory optimization.
\newblock In {\em 2012 IEEE/RSJ International Conference on Intelligent Robots
  and Systems}, pages 4906--4913. IEEE, 2012.

\bibitem{uehara2021pessimistic}
Masatoshi Uehara and Wen Sun.
\newblock Pessimistic model-based offline reinforcement learning under partial
  coverage.
\newblock {\em arXiv preprint arXiv:2107.06226}, 2021.

\bibitem{xu2021error}
Tian Xu, Ziniu Li, and Yang Yu.
\newblock Error bounds of imitating policies and environments for reinforcement
  learning.
\newblock {\em IEEE Transactions on Pattern Analysis and Machine Intelligence},
  44(10):6968--6980, 2021.

\bibitem{combo}
Tianhe Yu, Aviral Kumar, Rafael Rafailov, Aravind Rajeswaran, Sergey Levine,
  and Chelsea Finn.
\newblock Combo: Conservative offline model-based policy optimization.
\newblock In M.~Ranzato, A.~Beygelzimer, Y.~Dauphin, P.S. Liang, and J.~Wortman
  Vaughan, editors, {\em Advances in Neural Information Processing Systems},
  volume~34, pages 28954--28967. Curran Associates, Inc., 2021.

\bibitem{mopo}
Tianhe Yu, Garrett Thomas, Lantao Yu, Stefano Ermon, James~Y Zou, Sergey
  Levine, Chelsea Finn, and Tengyu Ma.
\newblock Mopo: Model-based offline policy optimization.
\newblock In H.~Larochelle, M.~Ranzato, R.~Hadsell, M.F. Balcan, and H.~Lin,
  editors, {\em Advances in Neural Information Processing Systems}, volume~33,
  pages 14129--14142. Curran Associates, Inc., 2020.

\end{thebibliography}

\end{document}